\documentclass[letterpaper, 10 pt, conference]{ieeeconf}  

\IEEEoverridecommandlockouts                              
\usepackage{graphicx} 
\usepackage{epsfig} 
\usepackage{epstopdf} 
\usepackage{mathptmx} 
\usepackage{times} 
\usepackage{amsmath} %
\usepackage{amssymb}  
\usepackage{url}
\usepackage{siunitx}
\usepackage[USenglish,american]{babel}
\usepackage{kantlipsum} 
\usepackage{mwe} 
\usepackage{array} 
\usepackage{algorithm}		
\usepackage{algorithmic}	
\usepackage[percent]{overpic}	
\usepackage{xcolor}	
\usepackage{tikz}
\usepackage{float}
\usepackage{placeins}

\newcommand{\Function}[1]{\ensuremath{{\,\textsc{#1}}}}

\title{\LARGE \bf
Baxter's Homunculus: Virtual Reality Spaces for Teleoperation in Manufacturing
}

\author{Jeffrey I Lipton$^{1}$, Aidan J Fay$^{1}$, and Daniela Rus$^{1}$
\thanks{$^{1}$Computer Science and Artificial Intelligence Lab, 32 Vassar Street, Cambridge, MA, 02139, USA
        {\tt\small jlipton at mit.edu}}
}

\begin{document}

\maketitle
\thispagestyle{empty}
\pagestyle{empty}

\begin{abstract}

Expensive specialized systems have hampered development of telerobotic systems for manufacturing systems. In this paper we demonstrate a telerobotic system which can reduce the cost of such system by leveraging commercial virtual reality(VR) technology and integrating it with existing robotics control software. The system runs on a commercial gaming engine using off the shelf VR hardware.  This system can be deployed on multiple network architectures from a wired local network to a wireless network connection over the Internet. The system is based on the homunculus model of mind wherein we embed the user in a virtual reality control room. The control room allows for multiple sensor display, dynamic mapping between the user and robot, does not require the production of duals for the robot, or its environment. The control room is mapped to a space inside the robot to provide a sense of co-location within the robot. We compared our system with state of the art automation algorithms for assembly tasks, showing a 100\% success rate for our system compared with a 66\% success rate for automated systems. We demonstrate that our system can be used for pick and place, assembly, and manufacturing tasks.   

\end{abstract}

\section{INTRODUCTION}
We believe that the future of manufacturing will require a combination of robots and people, interacting in the physical and virtual worlds to allow humans and robots to enhance each other's capabilities. Humans could supervise the robots from a distance and, when necessary, connect seamlessly to the robot’s task space. There they could control the robot to do the task, teach the robot to do a new task, or help the robot to recover from a failure or an unexpected situation. We demonstrate a system for achieving such human-robot collaboration using the homunculus model of mind. The homunculus is a logical fallacy illustrated in the "Cartesian Theater" criticism of Descartes's mind body dualism\cite{gregory1987oxford}\cite{dennett1992time}. It is said to be the thing sitting in a control room inside of a persons head, looking through a person's eyes, controlling their actions. It is a non-terminating recursive definition because inside each homunculus it is implied there is another smaller homunculus. While this is a terrible definition of human intelligence, it is an appropriate definition of the intelligence of a teleoperated robot. Inside the robot there is a human which is aware, in a control room, seeing through its eyes and controlling its actions.  Our homunculus-inspired approach enables the human user to have the same view point and feel co-located with the robot, without the need for the user and the robot to have tightly coupled states. 

Unlike current and past attempts to apply VR systems with teleoperation, our system uses a virtual reality control room (VRCR) to decouple the user's and robot's inputs and outputs and provides an adjustable mapping between them. This VRCR is mapped to a space inside of the robotic system space. Our system uses current consumer virtual reality (VR) systems, allowing us to leverage existing commercial solutions along with gaming infrastructure to perform manufacturing tasks. We providing the user's senses stimuli from a local machine, while remote operating the robotic system. This has allowed us to run over a local wired network, a wireless connection in the same building, and control a robot over a hotel's wireless connection from a different city.

We tested our system by having it pick an item and stack it for assembly with greater accuracy and comparable speed to state of the art In-hand Object Localization (IOL) systems \cite{Changhyun}. Our system had a 100\% success rate with an average time of 52 seconds in 20 trials, compared with a 66\% success rate reported for the state of the art algorithm. We also used the system to perform fixture-less assembly tasks, and to manipulate and secure wires onto wooden structures with a commercial staple gun.  The system can pick up screws, handle flexible materials, and complex shapes by leveraging the VR infrastructure and human intelligence.

In this paper we contribute:
\begin{itemize}
\item A method for applying the homunculus model of the mind for teleoperation systems to enable user and robot virtual co-location
\item A new virtual reality based telerobotic architecture for control of robotic systems
\item A series of demonstrations of the telerobotic system with comparisons to current automation techniques.
\end{itemize}

\section{BACKGROUND}
\begin{figure*}[!htbp]
\begin{centering}
\includegraphics[width=0.98\textwidth]{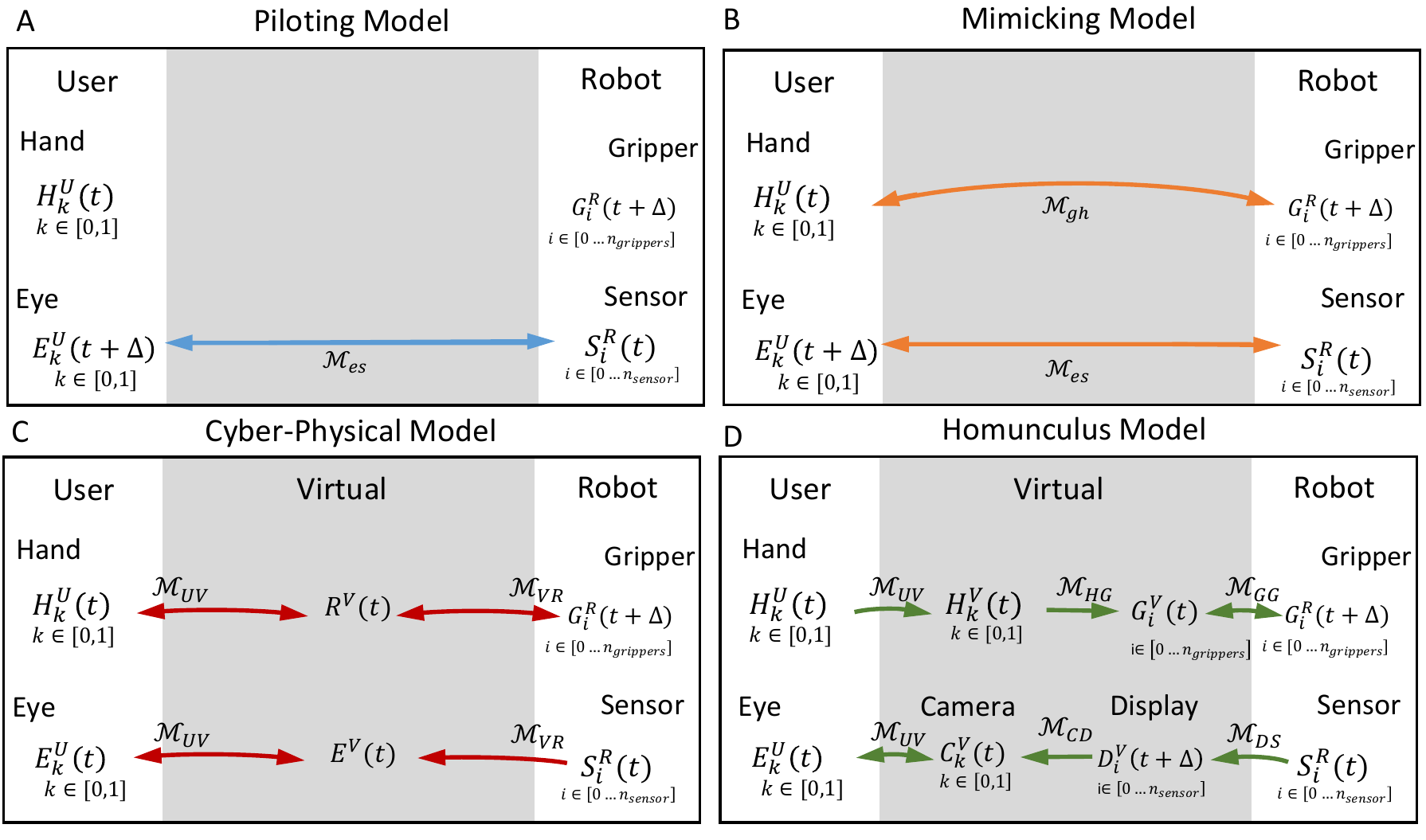}
\caption{Telerobotic System's Mapping Models. Each class of system maps different amounts of information from the robotic space and user space into each other, or into a virtual space. Mapping direction is denoted by an arrow and the mapping is named by $\mathcal{M}$ with subscripts. The piloting and mimicking models can have significant delays or have limiting mappings. The Cyber-Physical model introduced either a separate virtual space or have the user interact directly with a dual of the robot and the environment. Our Homunculus model combines the best of all three other methods. The virtual space decouples the human from the robot, and the mappings for sensors and grippers need not be identical. This allows the user to feel as if they inhabit the space inside the robot without the discomfort of the piloting and mimicking models, transmit less data, and have more information available to them. We also allow the mapping $\mathcal{M}_{HG}$ to be changed by the user as needed 
\label{fg:mappings}}
\end{centering}
\end{figure*}
Teleoperation systems map objects between the user's space $U$ to the robot's space $R$. The state of the user's hands ($H^U_k$) and eyes ($E^U_k$) are affected by and effect a robot's grippers ($G^R_i$) and sensors ($S^R_i$) using a series of mappings($\mathcal{M}$) shown in Figure \ref{fg:mappings}.  The user and robot may have completely different degrees of freedom, ranges of joints and number of states, making mapping methods critical to system function\cite{wang2015shared}. We can model each item as having a position $\overrightarrow{r}$, and orientation $(\hat{n},\theta)$ with super script denoting reference frame, first subscript referencing the object (H,E,S, or G) and the second subscript referencing the item index.

 Various teleoperation systems have different mappings and required latencies between action and feedback. Engaging a person's proprioception is what causes a person to feel present in a virtual reality environment \cite{mine1997moving}. However, unexpected incongruities, such as delays or relative motions, between proprioception and vision can lead to the nausea and motion sickness known as simulator sickness.  \cite{kennedy1996postural}. This means quickly mapping user eye movement and hand movement to an new stimulus is critical to user experience.
 
In piloting systems, a user operates a standard control such as a keyboard and mouse, haptic pen, or video game controller \cite{stilman2008humanoid}\cite{rodehutskors2015intuitive}\cite{wang2015shared}. As seen in Figure \ref{fg:mappings}A, We can model  $\mathcal{M}_{es}$ as simply delivering camera data to a display without receiving any feedback from the user states.  
We can model the mapping from eyes to sensors (  $\mathcal{M}_{es}$ ) as : 
\begin{equation} \label{eq:mes:r}
\overrightarrow{r}^R_{S,i}(t+\Delta_r) = \bar{C}_r \bar{T}_{RU} \overrightarrow{r}^U_{E,k}(t)\delta_{ik} + \overrightarrow{O}^U_{E,k}
\end{equation}
\begin{equation} \label{eq:mes:n}
\hat{n}^R_{S,i}(t+\Delta_n) = \bar{C}_n \bar{T}_{RU} \hat{n}^U_{E,k}(t)\delta_{ik}
\end{equation}
\begin{equation} \label{eq:mes:theta}
\theta^R_{S,i}(t+\Delta_n) = \bar{C}_{\theta}\theta^U_{E,k}(t)\delta_{ik} 
\end{equation}
$\bar{T}_{RU}$ is the transform from the U system to the R system of the form $\bar{T}_{RU} = \lambda_{RU}\bar{R}_{RU}$, where $\lambda_{RU}$ is a scale factor and $\bar{R}_{RU}$ is the rotation of the coordinate systems. $\Delta$ represents delays in the system,$\delta$ is the Kronecker delta, and $\bar{C}$ represents the coupling of the system's component's states as either $\bar{0}$ or $\bar{I}$. 

Telepresence systems can give an immersive experience by using a mimicking model. In these systems there is a direct connection between a user's state and the robots as seen in Figure \ref{fg:mappings}B. The user is meant to feel co-located with the robot, by closely mapping their states to each other. If the cameras are stationary relative to the robot then  $\bar{C}_r, \bar{C}_n, $ and $ \bar{C}_\theta$  are  $\bar{0}$, and the mapping $\mathcal{M}_{es}$ is only valid if one holds one's head perfectly still.  The user's mind constantly expects motion and does not receive it because of the bad mapping. This can lead to user fatigue and nausea.

An alternative is to have the mapping $\mathcal{M}_{es}$ include movement feedback. If only orientation feedback is used, the mapping does not compensate for non-rotating movements of the users head, and $ \bar{C}_n$, and $\bar{C}_\theta$ are $\bar{I}$. This requires the user to maintain position for the mapping to be valid. The UPenn developed Dora platform attempts to mimic the position and orientation of the user's eyes with a moving camera system \cite{Spectrum}. The large and complex nature of the camera rig limits its viability. Additionally, a hardware-in-the-loop solution will have to eliminate the delays from the network transmission and hardware response since it will need to update the user with appropriate stimuli every 90th of a second. 

Mimicking the body movements of a user can also be problematic. To provide a complete measurement of the states $H^U_k$ the body movements can be captured using hand controllers, Kinect sensors, Vicon systems, or robotic arm systems \cite{HydraRift}\cite{fritsche2015first}\cite{reddivari2014teleoperation}\cite{kofman2005teleoperation}. The incongruities between a user's body shape and the robots can be limiting. $\mathcal{M}_{gh}$ is of a similar form to equations \ref{eq:mes:r}, \ref{eq:mes:n}, and \ref{eq:mes:theta}. Robot arms can have different numbers of joints, degrees of freedom, sizes, and ranges from human arms. Mapping movements directly can lead to user discomfort or failure to utilize the entire work-space.  Baxter for example can reach positions and contortions that are painful or impossible for a user, such as putting both arms behind itself.  One solution for these systems is to closely mirror the abilities of a human user by changing the shape and design of the robot. The JUSTIN system, for example, is designed to closely mimic human form \cite{kremer2009multimodal}. However this level of mirroring is not compatible with a wide range of common cheap robots, and limits the system applicability. 

A third approach is the Cyber-Physical System (CPS) approach(See Figure \ref{fg:mappings}. In these systems a virtual dual of the robotic system($R^V$) and environment($E^V$) and dual of the user is represented in a computer \cite{passenberg2010survey}. The mappings $\mathcal{M}_{UV}$ from the user space and $\mathcal{M}_{VR}$ from the robot space are similar. 

The user receives stimuli from the dual model of the robot and environment, as the duals are constantly updated from the actual robot and environment. This requires making a complete model of the robot and environment, and transmitting updates to the models over the network. A version of this has used to generate a shared virtual space by embedding the dual of the robot environment and user in a globally shared virtual space \cite{guerin2014adjutant}. This relaxes the requirements on the mappings for the user and the robot states. The users see themselves as separate from the arm but both are inhabiting a shared virtual space. This is a great tool for training robotic systems and teaching repetitive tasks, it lacks the human in the loop of the controls, making it difficult to respond to constantly changing situations\cite{guerin2014adjutant}. It also requires a large amount of data to capture the environment for virtual representation, or a significant amount of a priori knowledge for building models of the environment.

\section{Design of VR Environments for The Homunculus Model}
\begin{figure*}[!htbp]
\begin{centering}
\includegraphics[width=0.98\textwidth]{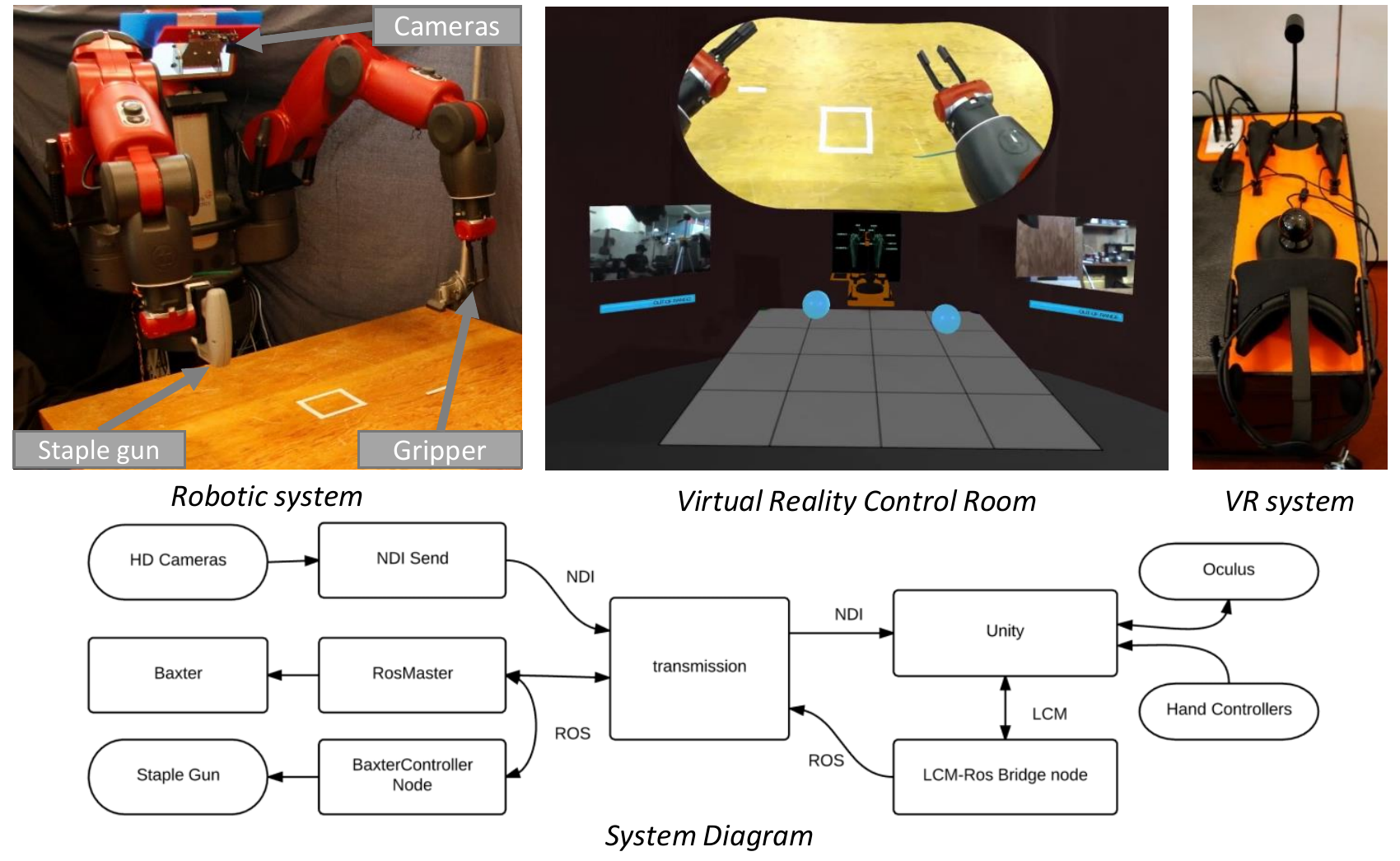}
\caption{Telerobotic system for Homunculus Model. A Baxter robot is outfitted with a stereo camera rig and various end effectors. In Unity, a virtual control room is established which uses an Oculus Rift CV1 headset and Razer Hydra hand trackers for inputs and displays. This allows the User to feel as if they are inside Baxter's head while operating it. Displays from each hand's sensors can be integrated into the control center. Only information from the cameras and commands, and state information need to be sent over the network. The transmission component can be non-existent, or a Canopy system for internetwork communications. The ROS nodes can be on the same machine, or different machines. 3D reconstruction from the stereo HD cameras are done by the humans visual cortex rather than by a GPU or CPU. 
\label{fg:system}}
\end{centering}
\end{figure*}

\begin{figure}[!htb]
\begin{centering}
\includegraphics[width=0.98\columnwidth]{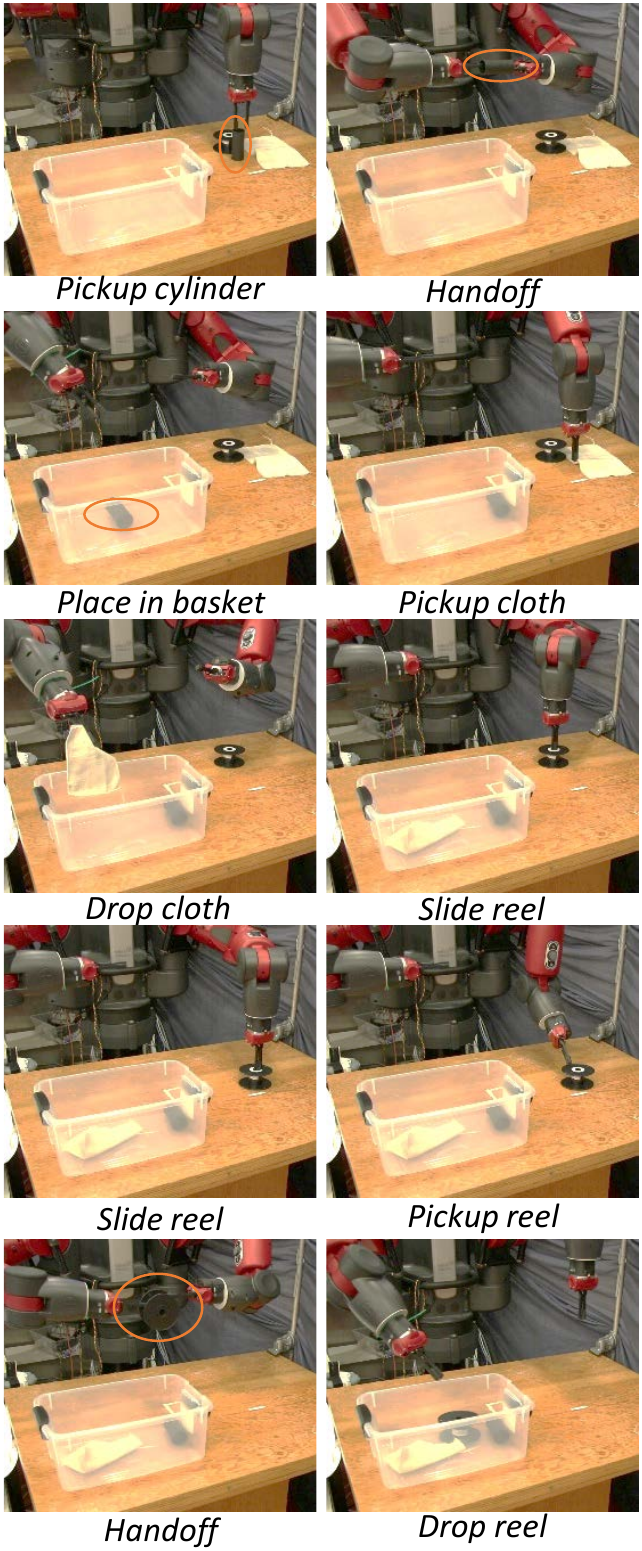}
\caption{Manipulating Various Shapes and Flexibilities. Users often have very astute perception and knowledge of an object's shape material properties. In this example a user is able to pick up a cylinder, a piece of cloth and a reel, hand them to the other hand, and place them in a bucket. The user needs to grasp from multiple angles, drag objects, and coordinate transfer to accomplish the task. In hard to see images, the item is circled in orange.}
\label{fg:binning}
\end{centering}
\end{figure}

\begin{figure*}[!htbp]
\begin{centering}
\includegraphics[width=0.98\textwidth]{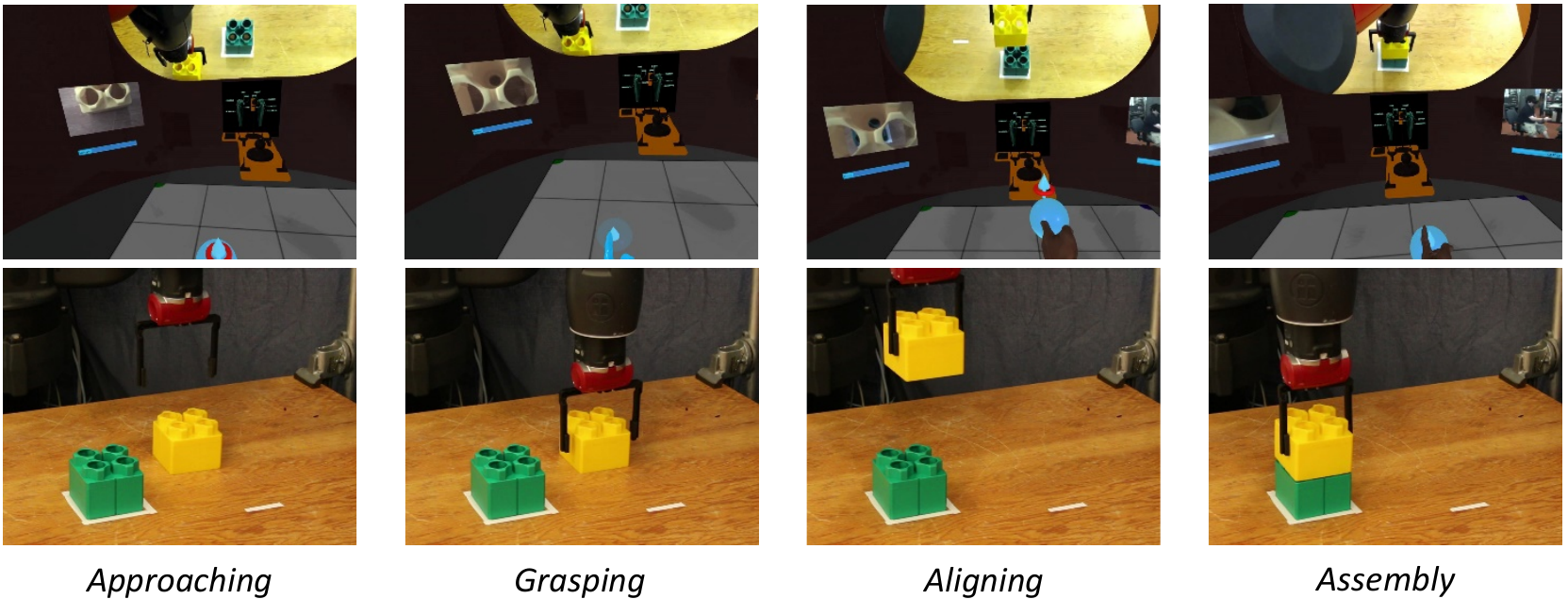}
\caption{Block Assembly. A series of blocks are grasped and stacked as a test of the system performance. A User can use the stereo cameras and hand cameras at the same time to approach, grasp, align and then assemble two blocks. }
\label{fg:assembly}
\end{centering}
\end{figure*}

In the homunculus model, a virtual reality control room (VRCR) provides a means of displaying information, presenting controls, while feeling present inside the robot(see Figure \ref{fg:system}). The human user space is mapped into the virtual space, and the virtual space is then mapped into the robot space. 
The virtual cameras allow the system to compensate for head movement. Since the VRCR is rendered on a local computer, in a separate loop from the robot movements and sensor feeds, it is possible to rapidly update the users stimuli. VR applications have a minimum frame rate of $60 Hz$ to prevent simulator sickness \cite{borg2013using}. Using commercial VR game engines to render most of the environment allows for the use of previously developed solutions such as Asynchronous Timewarp \cite{TimeWarp}. 

Unlike the CPS model the space is not shared with a virtual dual of the robot, nor does the user interact with a complete dual of the robot. Instead the user interacts with controls and markers that represent the last known position and orientation of the robot. This decouples the mapping $\mathcal{M}_{VR}$ into $\mathcal{M}_{GG}$ and $\mathcal{M}_{DS}$. $\mathcal{M}_{GG}$ is of the form: 
\begin{equation} \label{eq:mgg:r}
\overrightarrow{r}^V_{G,k}(t) = \bar{T}_{GG} \overrightarrow{r}^R_{G,k}(t) + \overrightarrow{O}^R_{G,k}
\end{equation}
\begin{equation} \label{eq:mgg:n}
\hat{n}^V_{G,k}(t) = \bar{T}_{GG} \hat{n}^R_{G,k}(t)
\end{equation}
\begin{equation} \label{eq:mgg:theta}
\theta^V_{G,k}(t) = \theta^R_{G,k}(t)
\end{equation}
$\bar{T}_{GG}$ is of the form $\lambda_{GG}\bar{R}_{GG}$ where $\lambda_{GG}$ is a scale factor and $\bar{R}_{GG}$  is the rotation of the coordinate systems. $\mathcal{M}_{DS}$ is of a similar form but with $\bar{T}_{DS} = \lambda_{DS}\bar{R}_{DS}$. We set $\lambda_{DS} >> \lambda_{GG}$ due to the parallax of the 3D camera system and its scaling, and the scaling in the virtual space. This allows the user to see the objects in the window as much larger than themselves .
the mapping $\mathcal{M}_{HG}$ is of the form: 
\begin{equation} \label{eq:mhg:r}
\overrightarrow{r}^V_{G,k}(t) = \bar{C}_{r} \overrightarrow{r}^V_{H,k}(t)
\end{equation}
\begin{equation} \label{eq:mhg:n}
\hat{n}^V_{G,k}(t) = \bar{C}_n \hat{n}^R_{H,k}(t)
\end{equation}
\begin{equation} \label{eq:mhg:theta}
\theta^V_{G,k}(t) = \bar{C}_{\theta}\theta^V_{H,k}(t)
\end{equation}
This allows the hands to selectively engage their connection to the grippers for position and orientation. The user is then free to move, fidget, or change their focus without effecting the state of the robot. The position and orientation of the robot arms' end-effectors are represented in the virtual space by objects which can be manipulated by the user to give commands to the robot. In Algorithm \ref{algo:Arm}, we see that the user can select the mapping $\mathcal{M}_{HG}$ they will use. Once selected, the user can grab the virtual gripper object to couple their movements to the robots. When released, the system uses a motion planning solver to find the joint states. If a solution is not found, the object is turned to an error state. If the virtual gripper is not coupled, the user's position and actions do not effect the robots arm state. 

Rather than reconstruct the dual of the physical environment, we can provide stereo views from which the user's brain can extract the 3D data. This reduces the amount of data which must be streamed and processed by the system. The VRCR is modeled after the head of the robot with a window displaying a stereo camera feed of the robot's work space.  The virtual cameras can float and respond to movement, while the virtual window displays the physical cameras' data. Unlike the mimicking model many different sensors can be displayed in a VRCR without overlaying on top of other camera and sensor feeds.  Multiple types and sources of information can be arranged in a virtual space.

Together these abilities represent a powerful tool for interfacing with a robot. A User is capable of selectively changing the focus of their attention between the whole system and a single arm by shifting their body. A user can turn towards a single side of the room and focus on a single arm's camera and range finder, while maintaining sight of the arm in 3D with their peripheral vision.  A user can plan movements based on the relative distance between the arm's current location marker and their hand while looking at the live display of the arm. Relative positioning of the arms to each other can be achieved by seeing the current location markers and commanded end points of both arms at the same time. 

\begin{algorithm}[h!]
\caption{Control of Arm }
\label{algo:Arm}
\begin{algorithmic}[1]
\STATE User selects type of model $\mathcal{M}_{HG}$ by positioning fingers
\STATE User grabs virtual gripper $G^V_i$ with hand $H^V_k$
\IF{\Function{Type}($\mathcal{M}_{HG}$) is position and orientation} 
\STATE {$\bar{C}_r \leftarrow I_3 , \bar{C}_n \leftarrow I_3 ,\bar{C}_{\theta} \leftarrow I_1$} 
\ELSIF{\Function{Type}($\mathcal{M}_{HG}$) is position only} 
\STATE {$\bar{C}_r \leftarrow I_3, \bar{C}_n \leftarrow \bar{O}_3, \bar{C}_{\theta} \leftarrow \bar{O}_1$}
\ENDIF
\STATE{User releases $G^V_i$}
\STATE{Joints solution $A \leftarrow \Function{Planner}(\overrightarrow{r}^R_{G,i},\hat{n}^R_{G,i},\theta^R_{G,i} )$}
\IF{$A$ is valid} 
\STATE {$\forall$ joints $J,  J_i \leftarrow A_i$}
\ELSE 
\STATE { \Function{Color}($G^V_i,error$) }
\ENDIF
\end{algorithmic}
\end{algorithm}

\section{The Homunculus Model System}
For our particular system we used an Oculus Rift as the VR headset and touch controllers, and a Baxter robot. Early in development Razer hydra controllers were also used for hand controllers. We chose the Baxter robot because it is commonly available, and though it looks humanoid, the range and DOF of the arms are distinctly not human like. Baxter provides a camera and range sensor in each arm. The Baxter robot was configured with either two parallel plate grippers or a gripper and a staple gun. Two Logitech C930e cameras were placed on Baxter's head as a stereo camera feed.

As seen in Figure \ref{fg:system} each arm's sensor information was displayed as a video screen and progress bar which indicated the range sensors value. These displays were on either side of the central window. This allowed a user to see through the hand camera, read the range sensor value, and see a stereoscopic view of an arm at the same time. 

To represent the control interface for the arm, we placed one glowing orb per arm in the space along with a marker for the arms current state. The orb is red if the robot was commanded to reach an unobtainable configuration, and blue if commanded to a valid position and orientation. The marker for each arm's current position and orientation was added to allow the user to compare changes in position. Additionally a virtual grid space was placed in the VRCR as a virtual reference surface. 

We allow the user to change the mapping $\mathcal{M}_{HG}$ to suit the task by changing how they held the controllers (See Algorithm \ref{algo:Arm}. Often a robot like Baxter keeps the grippers faced downwards towards a surface. This is an uncomfortable position for a human to hold. By selecting a different transform, the user can switch form having the gripper match the hands position and orientation, to simple matching position, with orientation fixed into the downwards position. This makes the system easier to use in pick and place operations. Using the touch system, if a user grabbed a control orb while their index finger was pointing out, the system would attempt to mimic the position and orientation of the user hand. If instead a user grabbed the control with their whole hand, the arm would only mimic the position, while maintaining a downward orientation.

Figure \ref{fg:system} shows the system diagram. Since the Oculus and other consumer VR head sets use the Unity framework, a Windows PC is needed for the user. Since Baxter uses ROS and Ubuntu, we needed a messaging system which could communicate between the systems in a fast manner. Lightweight Communications and Marshalling (LCM), developed by MIT and University of Michigan proved a sufficient solution \cite{huang2010lcm}. LCM however can only work over a small subnet,(IE on the same router). To enable internetwork operations, the Canopy framework was used to relay ROS massages between ROS master nodes on different networks\cite{Alexcannopy}. The Canopy framework subscribes to ROS topics, broadcasts them to server, and then transmits them to other ROS Master nodes that connect to the same server with the same group. This allows several ROS networks with separate master nodes to communicate to each other. This is represented in Figure \ref{fg:system} as the transmission element. The network with the User would convert LCM messages and ROS commands into each other for rebroadcast. For high definition video, the LCM/ROS infrastructure was insufficient. We used the NDI system by NewTek for transmitting and receiving a compressed HD video stream from the HD cameras. These streams were then integrated into Unity as a texture. Two planes, one displayed to each eye, have the camera feeds overlaid on them to provide a stereoscopic view out of the VRCR main window. 

\section{Results and Discussion }

\begin{figure}[!htb]
\begin{centering}
\includegraphics[width=0.98\columnwidth]{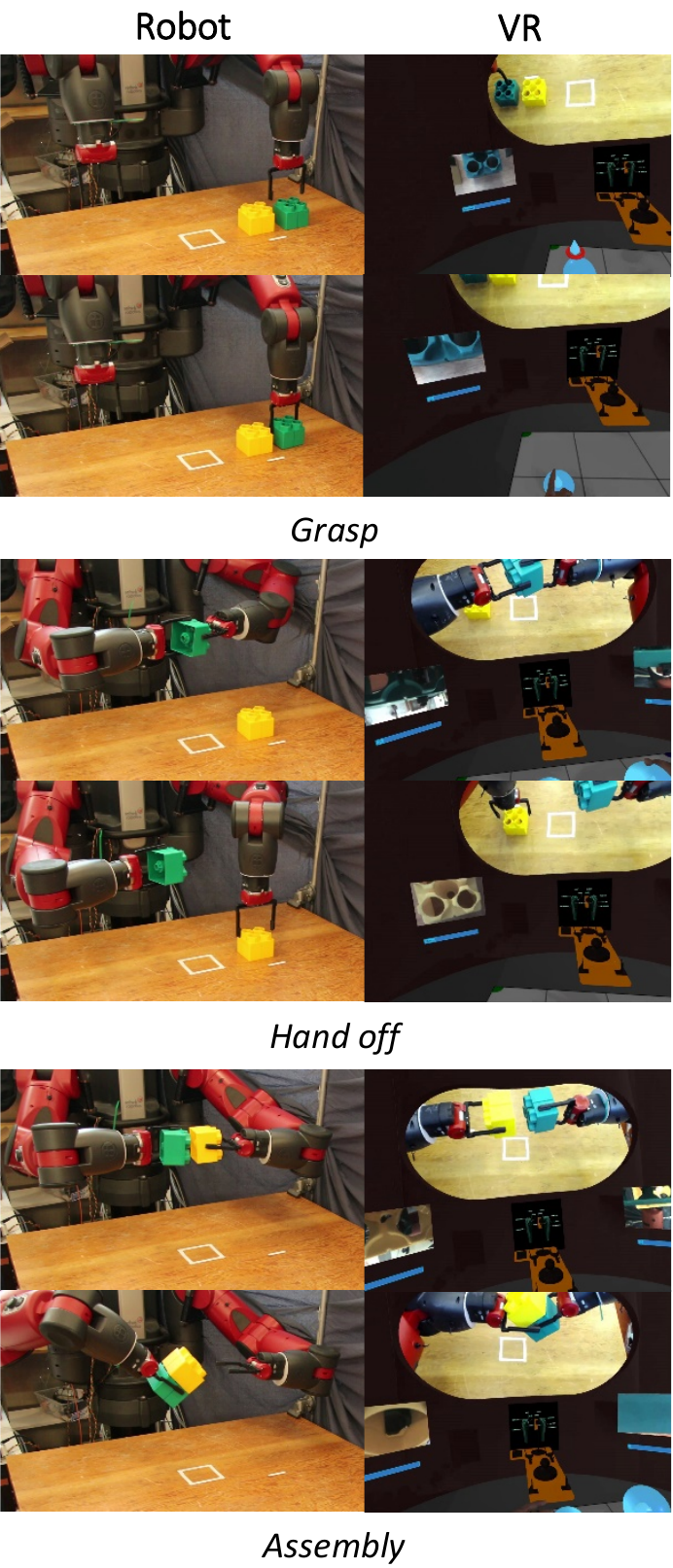}
\caption{Fixture-less Assembly by Teleoperation. A user is able to pick up a block, hand it off to the other hand, pick up a second block and stack them together. This requires the use of multiple $\mathcal{M}_{HG}$ to complete.}
\label{fg:fixtureless}
\end{centering}
\end{figure}

In order to test the system we devised a series of tasks for an expert user to perform. We focused on assembly, pick-and-place, and manufacturing tasks for the system. For pick-and-place operations, we tested the system's ability to handle objects of different shape and compliance. The user was tasked with picking up each item, transferring them from the left hand to the right hand, and then depositing them into a bin. This required the user to identify locations to grasp the object from, align the hands, re-grasp, and transfer to the bin. This is a non-trivial set of operations for a robotic system since it must be able to predict the movements of the cloth in response to grasping, understand the presence of the overhangs on the reel, and plan to drag the reel into a position where it can be grasped from the side. By contrast, a user is able to quickly execute these tasks in the teleoperation framework by leveraging their intuition for the objects' response to stimuli. It was useful to the user to be able to both visualize the overall space through the stereo window, and to be able to look through the hand camera to gauge the sense of alignment and distance to a target grasping location. In the supplementary video we see an expert user is able to use the hand cameras and virtual window to pick and place screws into a bin and lift and place a wire. Overall the system was able to handle many different types of items. We conducted evaluations with 8 different objects.  This mode of operation is made possible by the separation of $\mathcal{M}_{VR}$ into the $\mathcal{M}_{GG}$ and $\mathcal{M}_{DS}$ of the homunculus model. With the controls for the robot separated from a virtual representation of the robot, the multiple large displays can be placed further from the user in virtual space and the controls can be confined an appropriately mapped region of interest. 

\begin{table}[!htbp]
\setlength\tabcolsep{6pt}	
\renewcommand{\arraystretch}{1.5}	
\centering
\caption{The Success rate of assembly tasks relative to automated procedures with and without in-hand object localization (IOL)}
\label{model_table}
\begin{tabular}{cccc}
\hline
\multicolumn{1}{c}{Measured} & \begin{tabular}[c]{@{}c@{}}Automated\\  without IOL\end{tabular} & \begin{tabular}[c]{@{}c@{}}Automated \\ with IOL\end{tabular} & \begin{tabular}[c]{@{}c@{}}Teleoperated \\ VR System\end{tabular} \\ \hline
Successful grasping           & 100\%                                                            & 100\%                                                         & 100\%                                                             \\
Successful Assembly           & 41\%                                                             & 66\%                                                          & 100\%                                                             \\ \hline
\end{tabular}
\end{table}

Recently researchers at MIT had Baxter perform an assembly task using in-hand object localization (IOL) with a prime-sense sensor. Once an object was grasped in the hand, it was localized to improve the accuracy of an assembly task\cite{Changhyun}. We conducted teh same experiment using the Homonculus system (see Figure \ref{fg:assembly}) and measured performance over 20 trials.In table \ref{model_table}, we can see the success rate of using Baxter with a hard parallel plate gripper hand to grasp and assemble large blocks. While the state-of-the-art IOL algorithm can dramatically improve the success rate for the task, a human user had a perfect performance for a similar amount of execution time. A user took on average 52 seconds for the task with a 15 second standard deviation. In order to complete the task, it is useful to point Baxter's hand down towards the table. Our ability to vary $\mathcal{M}_{HG}$ aided the user in this task. By pressing a button the Razer, the $\mathcal{M}_{HG}$ would ignore the rotation information from the users hands, and only track position. This made grasping an object from above much more comfortable. A variant of this task is seen in Figure \ref{fg:fixtureless}. Here the user switches between two different $\mathcal{M}_{HG}$ to complete a fixture-less assembly task. When grasping items from the table, the rotation information is fixed, but when assembling in free space or handing the items off between hands, the rotation of the users hands are directly mapped to the robots. 

\begin{figure}[!htb]
\begin{centering}
\includegraphics[width=0.7425\columnwidth]{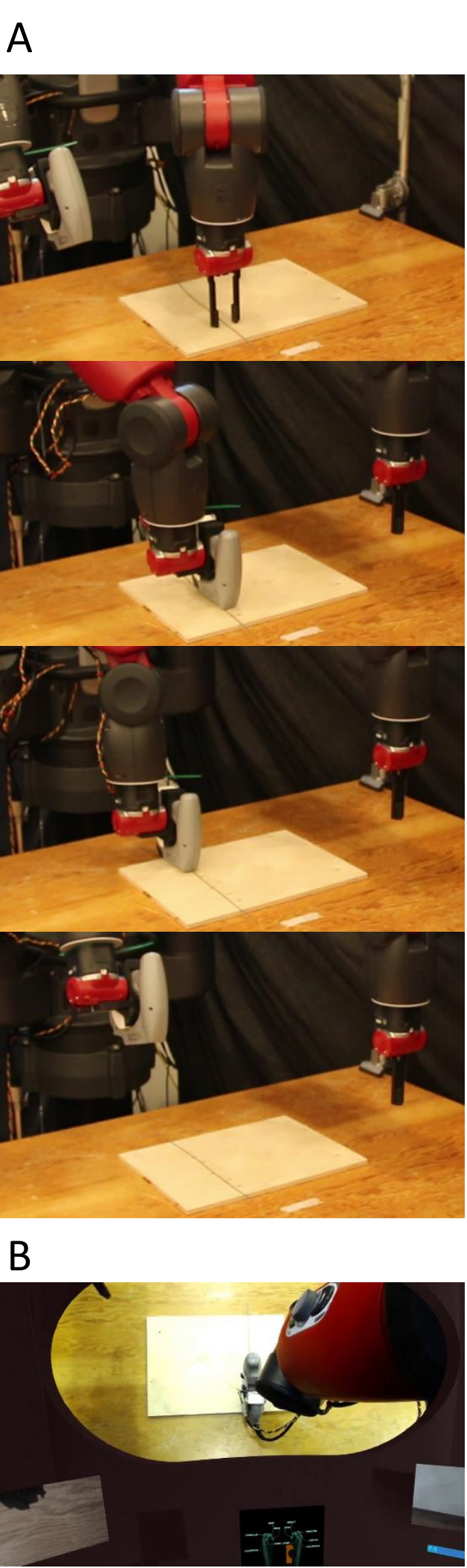}
\caption{Stapling wire  to a board. A user is able to place a thin wire with the gripper tool on the left arm and staple it in place with the right arm (A). From the virtual control room (B), the user can align the staple gun using their stereo vision.}
\label{fg:stapling}
\end{centering}
\end{figure}

For manufacturing a task, we had the user staple a wire to a board. We also tested the importance of the relative utility of the hand cameras. The user needed to be able to pick and place a wire onto a board with one arm, place a staple gun with the other, and fire the gun several times to lock the wire in place. As seen in Figure \ref{fg:stapling} the staple gun was attached to the side of the robot arm as an end-effector. This prevented the use of the hand camera when aligning the staple gun. The user was able to easily align with and pick up the wire while using the hand camera  Without the hand camera, alignment of the staple gun was inaccurate, and the user needed 5 attempts to secure 2 staples.  Despite this handicap, the user was able to use the 3D depth information from the virtual window to align the hand to the wire and staple the object. Future version of the system could include a pressure sensor and display to allow the user to know the force they are applying to the surface as well as a map of the pressure over the contact surface. This would allow the user to be sure the tool is lying evenly with enough force on the surface. 

In order to test systems ability over multiple network architectures, we controlled the robot in several different network configurations. The test above were done on a local network using a wired connection. We also controlled the robot in the same building over different networks via a wireless connection This required decreasing the update rate on the hand cameras to 5Hz. As a test of long range communication, the system was set up at the  Hyatt Regency Crystal City in Arlington, VA for the 2016 PI meeting of the National Robotics Initiative. There users were able to pilot a Baxter at MIT over the hotel's wireless Internet. This required limiting the frame rate on the hand cameras to 1Hz and the range update to 5Hz and reducing the resolution of the stereo cameras to 800x600. This was done to meet the 5 Mbps cap on the connection.

\section{CONCLUSION} 
In this paper we have developed a system for VR based telerobotics systems based on the homunculus model of mind. The system is highly flexible across network architectures and bandwidth allotments. By being able to rapidly change the mappings between human inputs and robot state and robot state to human inputs, we can improve utility and ergonomics. The ability of this architecture to leverage existing consumer grade hardware and software frameworks to remote operate existing robot systems provides it an ability to scale into wider deployment. Teleoperated robotic systems will allow humans the ability to work at scales and in environments which they cannot accomplish today. Barriers to working such as physical health, location, or security clearance could be reduced by decoupling physicality from manufacturing tasks. 

\section{ACKNOWLEDGMENT} 
We would like to thank the Boeing Corporation for their support for the project. We would also like to thank Changhyun Choi for his assistance with the Baxter system and for sharing his data on assembly. 

\bibliographystyle{IEEEtran}
\bibliography{refs}{}
\end{document}